# A Study on Knowledge Graph Embeddings and Graph Neural Networks for Web Of Things

Rohith Teja Mittakola
rohithtejam@gmail.com
CEA Paris-Saclay
Gif-sur-Yvette, France

Thomas Hassan
thomas.hassan@orange.com
Orange Labs
Cesson-Sévigné, France

## ABSTRACT

Graph data structures are widely used to store relational information between several entities. With data being generated worldwide on a large scale, we see a significant growth in the generation of knowledge graphs. *Thing in the future* is Orange's take on a knowledge graph in the domain of Web Of Things (WoT), where the main objective of the platform is to provide a digital representation of the physical world and enable cross-domain applications to be built upon this massive and highly connected graph of things. In this context, as the knowledge graph grows in size, it is prone to have noisy and messy data. In this paper, we explore state-of-the-art knowledge graph embedding (KGE) methods to learn numerical representations of the graph entities and subsequently, explore downstream tasks like link prediction, node classification, and triple classification. We also investigate Graph neural networks (GNN) alongside KGEs and compare their performance on the same downstream tasks. Our evaluation highlights the encouraging performance of both KGE and GNN-based methods on node classification, and the superiority of GNN approaches in the link prediction task. Overall we show that state-of-the-art approaches are relevant in a WoT context, and this preliminary work provides insights to implement and evaluate them in this context.

## 1 INTRODUCTION

Thing in the future [1] or Thing'in (in short) is a Web Of Things platform dedicated to reference connected and unconnected objects of the real world into a single graph, depicting their attributes and their interactions. This graph is built from different datasets depicting real-world data, where the nodes of the graph are the virtual counterpart of physical entities, i.e. their Digital Twin. The graph is built from diverse data sources, e.g. IoT platforms[2], opendata, research projects, etc. To enable the integration of data from a wide range of application domains, Thing'in's model is based on semantic web technologies: nodes and edges are mapped to ontologies and categorized with semantic attributes, forming a unique knowledge graph of *things*.

As this knowledge graph grows bigger, we observe problems like, noisy, messy or incomplete data (missing relations or unlabeled nodes). These problems are recurrent issues in the field of knowledge graph completion/refinement and curation, and are emphasized by the heterogeneous nature of its data sources. They can be mitigated with the help of machine learning models, however as large-scale IoT/WoT knowledge graphs are only emerging, applications of these methods in this context are scarce. Hence in this paper, we explored state-of-the-art machine learning techniques to refine the graph and mitigate data quality issues. In this paper, we are essentially interested in (i) Knowledge Graph Embeddings (KGE), and (ii) Graph Neural Networks (GNN).

Firstly, in order to apply machine learning techniques on non-euclidean graph data, it is required to learn numerical representation of its entities, i.e, low-dimensional vector embeddings. These embeddings capture the network information and graph properties which can then be used as input to several machine learning algorithms. Our first objective was to explore different knowledge graph embedding methods and learn embeddings for different sized sub-graphs collected from the Thing'in graph. In our context this is motivated by the fact that data in Thing'in does not always have explicit/ useful features for each entity in the graph. Subsequently, the embeddings were used on formulating graph downstream tasks like node classification, link prediction and triple classification, using standard classification algorithms (e.g. Support Vector Machines). Our second objective was to build graph neural networks to handle different graph related tasks and also compare their performance with methods involving KGE.

In particular, we are interested in the integration of graph neural networks and knowledge graph embeddings, and the impact of this integration on downstream performance. Graph Neural Networks require initial node representations: the numerical representations learned from Knowledge Graph Embeddings were used as input to GNNs, and evaluated against other representations such as degree, coloring number (which are graph centrality-based features) and other graph embedding methods like DeepWalk [18]. Then different GNNs models were used to formulate the same downstream tasks to compare their performance between them and with methods using KGE. Additionnally, to account for the multi-relational aspect of the knowledge graphs, GNN based models like R-GCN [22] were also explored.

Rest of this paper is divided in 5 sections. Section 2 discusses related work and most relevant state-of-the-art approaches for knowledge graph refinement. Section 3 and 4 describes respectively the datasets used in our evaluation, and detail the different tasks tackled. Section 5 explains the experimental setup and evaluation protocol for each task. Finally, section 6 provides a conclusion on experiments results and give insights for future work.

## 2 RELATED WORK

Formally, a knowledge graph is written as $\mathcal{G} = \{(h, r, t)\} \subseteq \mathcal{E}$ x $\mathcal{R}$ x $\mathcal{E}$ where, $h, r, t$ are the head, relation and tail of the knowledge graph and $\mathcal{E}$ is the set of entities of $\mathcal{G}$ and $\mathcal{R}$ is the set of relations of $\mathcal{G}$. Embeddings or low dimensional vector representations can be defined as a mapping $f : v_i \longrightarrow y_i \in \mathbb{R}^d \; \forall i \in |\mathcal{G}|$ where, $y_i$ embedding is generated for all triples individually and $d$ is a hyperparameter which shows the number of dimensions of the embedding.

---
[1] https://hellofuture.orange.com/en/thingin-the-things-graph-platform/
[2] e.g. Orange's IoT platform Liveobjects https://liveobjects.orange-business.com/

The survey papers [5, 27] explain different types of **knowledge graph embedding** methods which can be broadly grouped into 3 types: translation, factorization and neural network based methods.

The rationale which separates the methods from each other is based on the type of loss function used by them and also the way they capture the knowledge graph patterns like Symmetry (reverse of a triple is also true: e.g., <X marriedTo Y>), Asymmetry (reverse of a triple is not true: e.g., <X childOf Y>) and some other properties like Inversion, Hierarchy, Composition. A good KGE method can capture all these properties as numerical representations which explains the structure and relations of the graph. In general, a knowledge graph embedding algorithm has the following parts: a scoring function (which takes triples as input and outputs a numerical value), a loss function, an optimizer and a strategy to generate negatives. Negatives or corrupted entities are the false cases of a triple (which do not belong in the graph) and they are used during the evaluation of a knowledge graph model.

Translation based models use distance based functions to generate the embeddings. There are multiple ways to compute the distance functions in the euclidean space and there exist different algorithms using such ideas [27]. If we consider a triple (h,r,t), the **TransE** [3] algorithm tries to find embeddings for all entities and relations such that the combination of head and a relation vectors results in the vector of the tail i.e, $h + r \approx t$. With this idea, the scoring function can be formulated as the distance (L1 or L2 norm) between $h + r$ and $t$.

$$f_r(h, t) = -||h + r - t|| \quad (1)$$

**TransR** [15] changes the way relation vectors are handled. TransE does not take into account the heterogenity of the entities and relations while TransR tries to fix it by separating entity and relation spaces. The entities (h and t) are vectorized in space $\mathbb{R}^d$ and the relation is represented in a relation specific space $\mathbb{R}^K$. Then the entities are projected into the relation space to calculate the scoring function. Further, more translation algorithms have been developed like TransD [12], TranSparse [13] and TransM [9] which follow the similar idea of translation with some changes in implementation.

**RESCAL** [17] algorithm is based on tensor factorization as multiple relations can be expressed easily with a higher order tensor. A three way tensor $\mathcal{X} \in \mathbb{R}^{n \times n \times m}$ ($n$ denotes the number of entities and $m$ denotes the number of relations), is formed where two modes are concatenated entities and the third mode holds the relations. The tensor holds value **1** denoting the presence of a relation between the entities and **0** if there is not any.

$$\mathcal{X}_k \approx AR_k A^T, for \; k = 1, ..., m \quad (2)$$

where $A \in \mathbb{R}^{n \times d}$ is a matrix which contains the embeddings of the entities. RESCAL uses rank-$d$ factorization to learn the embeddings of $d$ dimensions. The scoring function is defined as:

$$f_r(h, t) = h_T M_r t \quad (3)$$

where $h, t \in \mathbb{R}^d$ are embeddings of entities and $M_r \in \mathbb{R}^{d \times d}$ is the latent representation of relation. RESCAL is computationally expensive for larger graphs so the method DistMult [29] simplifies the complexity by restricting the $M_r$ matrix to a diagonal matrix. Building on this, Holographic Embeddings (HolE) [16] and Complex Embeddings (ComplEx) [24] tries to simplify the RESCAL idea and also improve it by incorporating new ideas.

The neural network based methods like Semantic Matching Energy (**SME**) [2] defines an energy function which is used to assign a value to the triple by using neural networks. The intuition used here extracts the important components from different pairs and then puts them in a space where they can be compared. For instance, consider the triple (h,r,t) and firstly each entity and relation is mapped to an embedding $E_h$, $E_r$ and $E_t \in \mathbb{R}^d$. Then the embeddings ($E_h$ and $E_r$) are used to find new transformed embeddings by using a parameterized function such as $g_h(E_h, E_r) = E_{hr}$. The same process is repeated for the embeddings $E_t$ and $E_r$ and their respective new transformed embeddings ($E_{tr}$) are learned. Now we define an energy function which takes these transformed embeddings as input and gives a value the triple as: $Energy, \mathcal{E} = f_r(h, t) = h(E_{lr}, E_{tr})$. **ConvE** [7], is the first model which applied convolutional neural networks in the context of knowledge graphs.

**Graph Neural Networks** use artificial deep neural networks to operate on graph data structures. The performance and expressiveness of GNNs has been impressive and extensive research is being done to improve GNN models [30]. Standard neural networks like convolutional neural networks and recurrent neural networks cannot be used directly on the graph data. GNNs use a technique called Neural Message Passing by which the information is propagated on the network based on the graph structure. The objective of a GNN is to learn the vector representation of the nodes such that it captures the neighbourhood information of each node. Given a graph $G(V, E)$ and its initial node attributes say, $X$, the neural message passing can be defined as:

$$h_v^{(0)} = x_v \quad (\forall v \in V) \quad (4)$$

$$a_v^{(k)} = f_{aggregate}^{(k)}(h_u^{k-1} | u \in \mathcal{N}(v)) \quad (\forall k \in [L], v \in V) \quad (5)$$

$$h_v^{(k)} = f_{update}^{(k)}(h_v^{(k-1)}, a_v^{(k)}) \quad (\forall k \in [L], v \in V) \quad (6)$$

where $f_{aggregate}^{(k)}$ and $f_{update}^{(k)}$ are parameterized functions and $h_u^{k-1}$ is the attribute of the node $u$ during the $k$-th neural message passing phase. $\mathcal{N}(v)$ is the neighbourhood of the node $v$ and all these neighbourhood nodes are given as input to the aggregation function [21]. In equation 4, $x_v$ is the node attribute and in equation (5, 6), $L$ are the number of layers in GNN. If the graph does not have a node attribute then, one hot encodings of node degree or other graph properties can be used. The propagation rule of a GNN can be generalized as:

$$h_v^{(t)} = \sum_{u \in \mathcal{N}(v)} f(x_v, x_{v,u}^e, x_u, h_u^{t-1}) \quad (7)$$

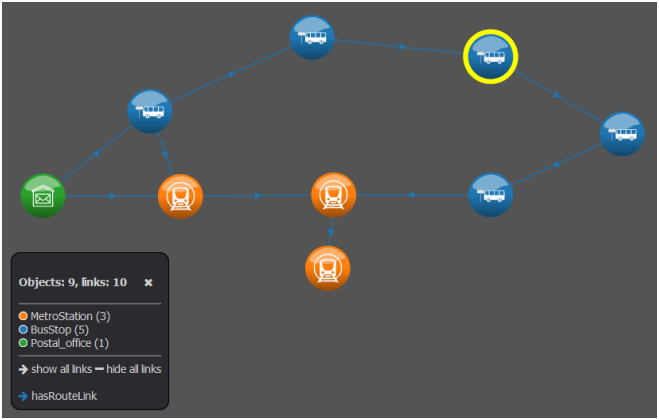

Figure 1: Example transportation graph

where $h_v^{(t)}$ is the node representation at time $t$ of node $v$, $x_{(v,u)}^e$ is the edge feature vector of edge $(v, u)$ and $h_u^{t-1}$ is the node representation of node $u$ in the previous time step. The convolutional neural networks, which are popular with image processing, can be used on graphs too and these models are getting popular in recent times. We can broadly divide the Graph Convolutional Neural Networks (GCNs) models into two types: spectral-based and spatial-based graph convolutional networks. For multi-relational graph modeling, we have methods such as Relational Graph Convolutional Networks (R-GCNs) [22].

## 3 DATA DESCRIPTION
### 3.1 Thing in the future graph

Data used in this paper is extracted from Thing'in the future Web of Things platform[3]. This platform hosts a graph consisting of millions of nodes and edges. The graph is a **labeled property graph** (LPG), modeled using the NGSI-LD[19] standard [4], however as a simplification it can be considered a knowledge graph of things. The objective of such a graph is to describe cyber-physical systems, i.e. the structural and semantic properties of environments such as cities, buildings, power-plants, etc.

All entities in the graph can be labeled with semantic properties[5], as well as "ad-hoc" properties and relations defined by the users, following the LPG paradigm. For this reason along with implementation concerns (see section 5.2), RDF embeddings methods such as Rdf2vec[20] were not used in this work, but are considered to be compared against in future work. Figure 1 shows an example piece of graph for transportation use case in a city, where nodes represent transport stations or post offices and hold properties such as id, geolocation, metadata, and edges represent physical connections, and hold information such as transport time or distance.

The graph is not fully connected, but different application domains can be connected through the graph, and interoperability is encouraged by the use of ontologies. As an example, a construction company can produce a digital description of a building using the IFC format. Translating and sharing this data in a common WoT platform, a facility maintenance operator can use the WoT platform to easily access desirable information for his use. However nothing ensures that building models are consistent accross the city. For instance, to describe the type/class of the building, one can use ontology classes **http://elite.polito.it/ontologies/dogont.owl#Building**, **https://w3id.org/bot#Building**, but also a user defined class from an ad-hoc schema. The same goes for any property or relationship between WoT objects. Even with ontology recommendations, nothing ensures that all application domains can be reconciled easily. For example the building model might be incompatible with nearby public transportation system, or power management. Additionally, allowing users to define their own schemas and labels inevitably leads to their proliferation, which cannot be reconciled using ontology mappings. This statement becomes bigger and bigger with scale, as more and more varied models have to inter-operate, and makes it unreasonable to query all systems at once, thus harder to build global services for use cases in logistics, industry 4.0, or smart city. That is why we aim to use machine-learning tasks (see section 4) to help reconcile models and mitigate the WoT graph inconsistencies, errors, noise, etc. **These are all critical task in WoT applications considering the number of IoT/WoT standards[11] and ontologies available**.

Table 1: Overview of Sub-graphs with basic information.

| Dataset | Description | Nodes | Edges | Labels |
|---|---|---|---|---|
| AC | Building data | 530 | 3142 | 10 |
| Adream | Building data | 778 | 3242 | 9 |
| C3 | Building data | 3581 | 8268 | 7 |
| EPFL | Building data | 67 | 140 | 6 |
| Garden | Building data | 299 | 2889 | 19 |
| Geonames | Points Of Interests | 11795 | 47158 | 2 |
| Meylan | City data | 5909 | 17704 | 38 |
| Poles | Telecom | 17506 | 35010 | 3 |

### 3.2 Sampled datasets

In this preliminary work we have sampled several sub-graphs that vary in size and application domains.

The graphs were collected in the *json* format from *ArangoDB*[6] and every graph takes the triples form of $\mathcal{G} = \{(h, r, t)\} \subseteq \mathcal{E}$ x $\mathcal{R}$ x $\mathcal{E}$ where every entity $\mathcal{E}$ has a label.

Table 1 shows the different datasets and their number of nodes, edges, and respective number of labels available for classification. It can be observed that the sub-graphs vary in number of entities, relations and labels. Each sub-graph belongs to a specific domain, representative of the domain heterogeneity in Thing in's LPG. One of our objectives is to detect if application domain has a significant impact on the performance of learning tasks.

Incoming stages of our research include experimenting with bigger datasets, using an adequate GPU-enabled hardware architecture. For confidentiality reasons, not all datasets could be shared. Available datasets are provided at [7]. The repository also include

---
[3] https://tech2.thinginthefuture.com
[4] https://en.wikipedia.org/wiki/NGSI-LD
[5] i.e. datatype properties and object properties, using various ontologies which are based on RDF https://www.w3.org/TR/rdf-schema/ or OWL https://www.w3.org/TR/owl2-syntax/ schemas
[6] https://www.arangodb.com/documentation/
[7] https://github.com/kgrl2021/submission-one

relevant implementation information e.g. hyperparameters for the experiments.

## 4 OVERVIEW OF GRAPH DOWNSTREAM TASKS

After we learn the embeddings, there are several graph downstream tasks which can be performed.

### 4.1 Node Classification

In the knowledge graph, the nodes have different attributes, including a label. In the knowledge graph as a whole, there are many erroneous/missing labels. These labels can be predicted by using graph embedding techniques and machine learning methods. This problem is termed as node classification problem. There are several applications of node classification in the context of Thing'in's graph:

- It can be used to predict an imprecise or missing node label. For example, Thing'in allows to model nodes with a high-level, abstract semantic label such as *owl:Thing*, or with a label defined by the user.
- In Thing'in, entities can be multi-labeled, thus node classification can be used to predict additional nodes labels for existing nodes.
- We can also detect outlier nodes that are labeled incorrectly.

***Task Formulation***. : Since the node labels are multiple (greater than 2) in most cases (see table 1), this problem can be formulated as a multi-class classification task. The knowledge graph embeddings can be directly used as input to machine learning algorithms like Support Vector Machines (SVM) or Random Forest to build a prediction model. Alternatively we can use Graph Neural Networks to tackle the same problem. Here, the input to the GNNs could be the learning based features, like knowledge graph embeddings, or centrality based node features like degree, graph coloring number etc [8]. We also tested GNN methods like R-GCN [22] which is then compared with the results from node classification task using KGE.

### 4.2 Link Prediction

It is the task of predicting missing edges or links in the graph and also predicting prospective edges (in case of a dynamic graph where the edges disappear and form based on the time-point of the graph). In the context of a knowledge graph this is often termed as knowledge graph completion. Recollecting that, a knowledge graph can be represented as a triple $G = (\mathcal{E}, \mathcal{R}) = \{(h, r, t)\}$, the knowledge graphs can be incomplete by either having a missing head or a tail. Using the facts in the knowledge graph our objective is to predict the unknown entities. It boils down to guessing a tail if it is a case of tail prediction $(h, r, ?)$ or guessing a head in case of a head prediction $(?, r, t)$.

***Task Formulation***: This task can be formulated in multiple ways. A straight forward way is to convert the problem into a binary classification task with two classes, namely the positive and negative classes (can be denoted as **1** for positive and **0** for negative). A standard machine learning model can be used to solve this binary classification problem and a prediction model can be obtained. For every link, we do a hadamard product of the two entities (source and target) to generate a single numerical representation which will be used as a feature in the binary classification task. We can also formulate the task as head prediction and tail prediction. In order to learn the knowledge graph embeddings and for model selection in specific, we used this case of entity prediction (head or tail). There are two different ways in which this task can be evaluated: Rank based evaluation[8] with metrics like Mean Reciprocal Rank (MRR) and Hits@K, and Threshold based evaluation[9] with metrics like average precision (AP) score. GNNs are used to formulate the task by splitting the edges into positive and negative edges (randomly sample negative edges). This task is then treated as a binary classification and a GNN model is built to predict the links and calculate metrics (AP). Multi-relational information of edges are used in case of R-GCN model.

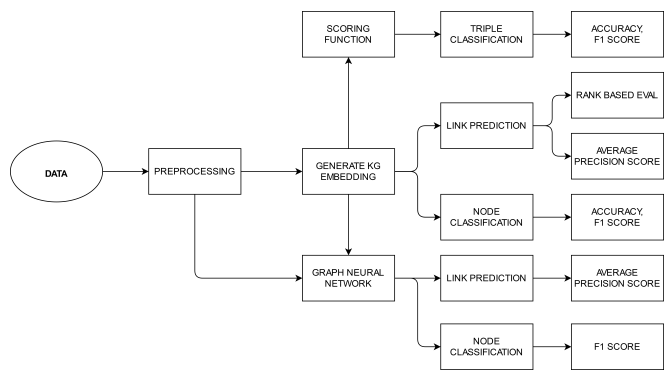

Figure 2: Flow chart of different methods and tasks

### 4.3 Triple Classification

Here, we aim to classify if a given triple is valid or not. The embeddings that are generated should be able to score the valid triples higher than the negative triples. Negative triples (or corrupted triples) can be generated by replacing the head or tail part of the triple with some random entity. This evaluation method works by setting a global threshold $\delta$. For instance, if a triple $(h, r, t)$ gets a score above $\delta$ then it is classified as positive or else negative. The threshold $\delta$ is a hyperparameter and it is adjusted according to the accuracy scores obtained on the validation set.

***Task Formulation***: This task boils down to a binary classification problem and we use scores (calculated when the embeddings are learned) of each triple as features to the machine learning model. The set of negative triples are labeled as **0** while, the positive triples are labeled as **1**. Machine Learning algorithms like SVM can then be used to handle the classification task.

## 5 EXPERIMENTS AND RESULTS

### 5.1 Experimental Setup

The experiments were performed on 8 different graph datasets belonging to different ontologies. Experiments include: generating knowledge graph embeddings, performing graph downstream tasks

---

[8]https://pykeen.readthedocs.io/en/latest/tutorial/understanding_evaluation.html
[9]https://scikit-learn.org/stable/modules/generated/sklearn.metrics.average_precision_score.html

A Study on Knowledge Graph Embeddings and Graph Neural Networks for Web Of Things

using both traditional machine learning methods and graph neural networks. All the experiments were performed on a 4-core CPU machine (2.1 GHz) with 16 GB RAM running Ubuntu OS and Anaconda Python 3.7 environment (GPU computation was not used). Firstly, the model selection for embedding generation was done based on the performance on link prediction task, and two types of evaluation methods were considered for this (Rank based and Threshold based evaluations). The embeddings, thus generated, were used as features in a node classification task and a machine learning model (SVM) was built to make predictions. During the embedding generation, each triple was given a score using a scoring function and these scores were used as input features for the triple classification task.

Graph Neural Networks were built to handle link prediction and node classification tasks. The initial node representations for these models were tested using two centrality based features (Degree, Graph Coloring Number) and two learning based features (Knowledge Graph Embeddings, DeepWalk [18]). Figure 2 shows the details of all experiments and tasks which are used in this paper.

## 5.2 Knowledge Graph Embeddings

The knowledge graph embeddings were generated using PyKEEN[10] python library and the hyperparameter tuning was done by Optuna[11] library. The hyperparameters like embedding dimension, learning rate, epochs, optimizers were tuned and popular knowledge graph embedding methods like TransE, TransR, DistMult and RotatE[23] were tested. The model selection was done based on the performance of embeddings on the Link Prediction task with threshold based evaluation. The embeddings generated using rank based evaluation showed relatively less performance on tasks like node classification, hence we preferred threshold based evaluation.

*5.2.1 Link Prediction.* : Table 2 shows the results obtained for the link prediction task using knowledge graph embeddings and both the evaluation methods. The triples were split into train and test in the ratio 90:10 respectively. Test size is chosen to be 10% as we wanted all entities present in test set also to be found in the train set (For some smaller graphs, this way of splitting was not possible if we chose a larger test set size). We included the metrics for both evaluation methods and training time for threshold based method. TransR [15] was found to be the best model to learn knowledge graph embeddings, thus all results reported in table 2 refer to TransR results. Figure 3 shows the embedding visualizations where each point in the figure represents an entity and the colour represents its class/ label. The visualization was made using t-SNE [25] which is a dimensionality reduction technique generally used to visualize high dimensional data.

Note: Since, we chose threshold based evaluation here, even for link prediction with GNNs, we used the average precision score to report the performance. Rank based metrics in table 2 are just for reference.

*5.2.2 Node Classification.* : Using the knowledge graph embeddings, we handled the task of node classification with the help of machine learning algorithms. Table 3 shows the accuracy and F1

[10]https://pykeen.readthedocs.io
[11]https://optuna.org

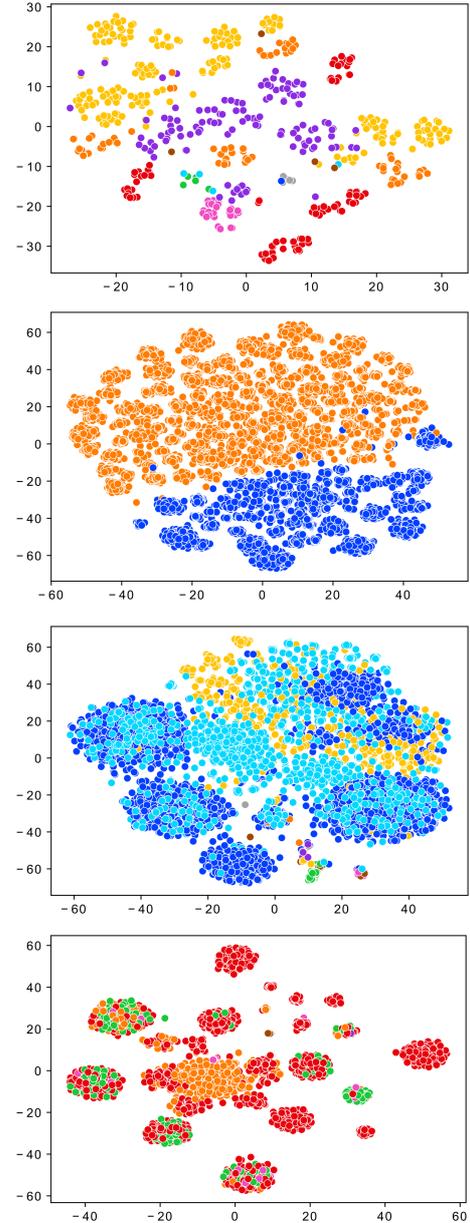

**Figure 3: t-SNE plot of embeddings of different sub-graphs (from top to bottom: *AC* building data, *Geonames* geolocation data, *Meylan* city data, *C3* building data)**

scores (in %) obtained for the node classification task. We used Support Vector Machine classifier and also Random Forest to perform the multi-class classification task. The test size was chosen to be 20% of the entire data and hyperparameters were tuned using 10-fold cross validation on the train data.

*5.2.3 Triple Classification.* : The experiments for triple classification were performed in two cases. Case 1: equal number of negative triples were generated to the existing positive triples and in case 2: twice the number of negative triples were generated. This was done to test the robustness of the machine learning algorithm in handling the imbalance of positive and negative instances. In both cases, the test size was chosen to be 20% of the entire data and hyperparameters were tuned using 10-fold cross validation on the train data. We used Logistic Regression and Support Vector Machine classifiers for this task and the scores from best model were shown in the table 4 (we found the best model to be SVM).

Table 2: Results of link prediction task using TransR method (Rank based metrics were included only as reference)

| Dataset | Rank-based | | | Threshold-based | |
|---|---|---|---|---|---|
|  | MRR | Hits@1 | Hits@10 | AP score | Time(Min) |
| AC | 47.9 | 35.5 | 70.0 | 79.7 | 66.4 |
| Adream | 41.1 | 31.0 | 57.9 | 81.4 | 2.3 |
| C3 | 18.7 | 10.7 | 35.6 | 79.3 | 37.1 |
| EPFL | 55.3 | 50.0 | 100 | 97.6 | 1.1 |
| Garden | 43.1 | 17.5 | 91.6 | 61.9 | 2.5 |
| Geonames | 81.3 | 74.3 | 95.1 | 93.1 | 21.7 |
| Meylan | 27.7 | 21.7 | 36.8 | 81.4 | 33.6 |
| Poles | 54.8 | 52.4 | 57.9 | 100 | 102.2 |

Table 3: Results of node classification task using KGE methods and R-GCN

| Dataset | SVM | | Random Forest | | R-GCN | |
|---|---|---|---|---|---|---|
|  | Accuracy | F1 Score | Accuracy | F1 Score | Accuracy | F1 Score |
| AC | 91.5 | 88.9 | 87.7 | 85.0 | 39.6 | 52.0 |
| Adream | 73.1 | 67.8 | 68.6 | 60.9 | 33.3 | 39.2 |
| C3 | 85.2 | 82.0 | 82.6 | 78.2 | 30.6 | 28.1 |
| EPFL | 78.6 | 69.1 | 78.6 | 69.1 | 57.1 | 62.3 |
| Garden | 61.7 | 56.3 | 70.0 | 64.7 | 33.3 | 46.6 |
| Geonames | 99.6 | 99.6 | 97.6 | 97.6 | 65.4 | 74.6 |
| Meylan | 89.6 | 88.7 | 85.6 | 84.0 | 50.9 | 60.9 |
| Poles | 100 | 100 | 100 | 100 | 61.4 | 62.1 |

5.2.4 *Discussion.* : The results from node classification (see table 2) task and also the triple classification (see table 3) task show that the knowledge graph embeddings were of good quality. Most of the datasets show an accuracy and F1 score above 80% in the node classification task, and above 90% in the triple classification task. The embeddings have captured the network structure and its properties which resulted in good performance. The poor performance on some datasets is presumably due to the fact that the graphs are too small in size (e.g. *EPFL*), which makes it difficult to learn good quality embeddings, or they have highly imbalanced class distribution (e.g. *Garden*, see table 1).

## 5.3 Graph Neural Networks

Initial numerical representations for every node in the graph are important to build a graph neural network model. Since we did not have any useful representations for our graph data, we experimented with different representation methods: Graph centrality based methods, i.e. In-degree and Coloring number, and Learning based methods, i.e. Knowledge graph embeddings, DeepWalk [18], and also, random embeddings. In total, we ran the experiments using 5 different node representation methods. The different GNN models we used in experiments include: GAT [26], GCN [14], ChebNet [6], GCN2 [4], GCN-ARMA [1], SGC [28], GraphSAGE [10] and R-GCN [22].

For each graph downstream task, different GNN models were tested and the factors behind picking the models include: general performance of individual algorithm, diversity of the set, and ease of implementation. Adding to this, we tested R-GCN [22] to use the multi-relational aspect of the graph which the other GNN methods do not account for. Because R-GCN generates its own embeddings the results were shown separately from other GNN methods. We wanted to perform graph analysis using GNNs with and without considering the multi-relational aspect of the knowledge graph, so we have two sets of analysis.

5.3.1 *Link Prediction.* :

Here, the problem was formulated as a binary classification[12] task and the results using GNN models are displayed in table 5 and the results using machine learning models like logistic regression and SVM are shown in table 6.

The GNN models were built using Deep Graph Library[13] (DGL) and out of all models we tested GCN and ChebNet performed better. The positive edges were split into 80% train and 20% test sets and equal number of negative edges were generated for each set.

Note: The best performing model and also the initial node representations were highlighted in the table for each graph data (see table 5). To model with the multi-relational aspect, we used R-GCN and tackled the same link prediction task. The results of R-GCN are shown in table 6 (results for *garden* dataset are not available).

---
[12]https://docs.dgl.ai/en/0.6.x/new-tutorial/4_link_predict.html
[13]https://www.dgl.ai

Table 4: Results for triple classification task (using SVM)

| Dataset | Case 1 | | Case 2 | |
| --- | --- | --- | --- | --- |
| | Accuracy | F1 Score | Accuracy | F1 Score |
| AC | 98.0 | 98.0 | 95.1 | 95.1 |
| Adream | 96.7 | 96.7 | 93.6 | 93.7 |
| C3 | 98.9 | 98.9 | 97.3 | 97.3 |
| EPFL | 67.9 | 67.0 | 61.0 | 46.2 |
| Garden | 95.6 | 95.6 | 93.6 | 93.7 |
| Geonames | 98.8 | 98.8 | 91.6 | 91.8 |
| Meylan | 96.1 | 96.0 | 92.9 | 93.0 |
| Poles | 97.6 | 97.6 | 96.4 | 96.4 |

Table 5: Average precision scores for different datasets using GNN models in link prediction task

| Dataset | Models | Initial Node Representations | | | | |
| --- | --- | --- | --- | --- | --- | --- |
| | | In-degree | Coloring Number | KG Embedding | DeepWalk | Random values |
| AC | GCN | 95.6 | 88.0 | 95.9 | 97.0 | 94.1 |
| | ChebNet | 92.2 | 88.6 | 90.2 | 91.2 | 87.2 |
| Adream | GCN | 94.2 | 89.1 | 95.8 | 93.9 | 93.1 |
| | ChebNet | 94.4 | 91.0 | 93.9 | 94.6 | 89.9 |
| C3 | GCN | 97.3 | 96.5 | 97.4 | 97.8 | 96.3 |
| | ChebNet | 96.4 | 96.5 | 98.0 | 98.2 | 95.7 |
| EPFL | GCN | 98.3 | 98.3 | 89.8 | 98.9 | 94.2 |
| | ChebNet | 100 | 99.9 | 98.9 | 100 | 99.9 |
| Garden | GCN | 88.4 | 77.7 | 93.1 | 91.3 | 90.9 |
| | ChebNet | 91.6 | 78.8 | 90.5 | 87.5 | 88.8 |
| Geonames | GCN | 97.6 | 98.5 | 97.7 | 97.9 | 97.5 |
| | ChebNet | 95.2 | 94.8 | 97.1 | 93.4 | 97.6 |
| Meylan | GCN | 96.2 | 61.7 | 98.0 | 96.1 | 95.2 |
| | ChebNet | 96.0 | 96.3 | 98.2 | 98.3 | 95.8 |
| Poles | GCN | 82.7 | 76.0 | 84.9 | 83.3 | 81.1 |
| | ChebNet | 80.9 | 81.3 | 83.6 | 84.3 | 81.7 |

Table 6: Average precision scores for different datasets in link prediction task using standard machine learning methods and R-GCN

| Dataset | Logistic Regression | SVM | R-GCN |
| --- | --- | --- | --- |
| AC | 52.2 | 80.7 | 90.4 |
| Adream | 56.9 | 89.2 | 92.1 |
| C3 | 68.9 | 96.2 | 97.4 |
| EPFL | 93.0 | 96.4 | 100 |
| Garden | 62.2 | 74.0 | NA |
| Geonames | 81.6 | 93.9 | 96.7 |
| Meylan | 59.9 | 95.7 | 97.0 |
| Poles | 68.6 | 97.1 | 87.4 |

*5.3.2 Node Classification.* : Here, we used Pytorch Geometric[14] (PyG) to build the GNN models. The table 7 shows the results with the 2 best performing GNN models and different initial node representations. The data was split into train, validation and test sets (in ratio 80:10:10). The hyper-parameter tuning was done with the help of Optuna python library using validation score and the final metrics were calculated based on the predictions on the test set. F1 score was chosen as metric since most of the graph datasets have highly imbalanced classes. To model the same problem using multi-relational information of edges, we used R-GCN with the same data and table 3 shows the results.

*5.3.3 Discussion.* : Both DGL and PyG implement several graph neural network models. PyG focuses more on node classification and included many examples for it. Hence, PyG was used in node classification and DGL for link prediction. In the link prediction task, we can see that GNN models (also, R-GCN) are superior to standard machine learning models as all graph datasets (except one, i.e, *Poles* data) show better performance with GNN models (see tables 5 and 6). We can infer that GNN's message passing technique was able to capture the network information well and majority of the datasets perform better when the initial node representations are learning based (for link prediction this observation is not well pronounced but for node classification, learning based features show better performance by a huge margin). This can be due to the fact that message passing further improves the quality of the learning based features (which have already captured some network properties). R-GCN was able to use the additional information of edge types and performed well on the link prediction task. However in the node classification task, R-GCN does not perform as well (see tables 6 and 3).

---
[14]https://pytorch-geometric.readthedocs.io

Table 7: F1 scores for different datasets using GNN models in node classification task

| Dataset | Models | Initial Node Representations | | | | |
|---|---|---|---|---|---|---|
| | | In-degree | Coloring Number | KG Embedding | DeepWalk | Random values |
| AC | ChebNet | 37.8 | 56.8 | 88.6 | 60.8 | 28.7 |
| | GCN2 | 57.9 | 56.8 | 98.2 | 65.8 | 37.4 |
| Adream | ChebNet | 55.7 | 58.2 | 67.9 | 70.6 | 29.0 |
| | GCN2 | 57.7 | 58.2 | 70.2 | 80.6 | 43.0 |
| C3 | ChebNet | 84.2 | 84.2 | 84.5 | 82.3 | 73.1 |
| | GCN2 | 84.2 | 84.2 | 86.2 | 80.8 | 84.2 |
| EPFL | ChebNet | 83.3 | 83.3 | 50.8 | 62.3 | 62.3 |
| | GCN2 | 83.3 | 83.3 | 90.5 | 42.9 | 83.3 |
| Garden | ChebNet | 40.8 | 46.2 | 67.9 | 49.7 | 61.0 |
| | GCN2 | 44.5 | 48.1 | 68.1 | 70.5 | 48.4 |
| Geonames | ChebNet | 77.6 | 78.1 | 97.8 | 99.8 | 67.0 |
| | GCN2 | 77.7 | 78.1 | 99.9 | 99.9 | 77.6 |
| Meylan | ChebNet | 70.0 | 69.7 | 85.5 | 79.9 | 66.6 |
| | GCN2 | 70.0 | 70.0 | 84.0 | 79.2 | 65.9 |
| Poles | ChebNet | 100 | 83.8 | 100 | 99.9 | 75.7 |
| | GCN2 | 93.1 | 83.8 | 100 | 100 | 83.8 |

## 6 CONCLUSION

In this paper, we studied the application of knowledge graph embeddings and graph neural networks in the context of a Web Of Things platform, *Thing in the future*. We have reviewed, selected and implemented relevant approaches that could respond to the objective of validating existing knowledge as well as inferring new knowledge in this type of knowledge graph. We built and evaluated different graph downstream tasks like node classification, link prediction and triple classification and all these tasks can be used to build several real life use-cases. The experiments show that graph neural networks can further enhance the performance notably for link prediction. One observation about the heterogeneous nature of the knowledge graph and our experiments is that by segmenting experiments for each sub-graph, some types of graphs (e.g. building data) tend to be smaller in size. This impacts the performance of algorithms negatively, as they require more data to train a good model, thus the model selection could indirectly depend on the nature of the dataset. The possibility of generalizing our results depend on the future research and experiments on a wide, diverse range of sub-graphs. Until then, we argue our current research stand as a good precursor to future investigations. We believe these preliminary results are encouraging and provide a good insight on building predictive machine learning models on data relating to WoT domain and help building some use-cases for information retrieval, but we need to investigate further with bigger datasets, and by combining datasets from different domains.

## Acknowledgement

We would like to acknowledge that the author, Rohith Teja Mittakola, was affiliated with Orange Labs, Cesson-Sévigné during the research for this paper.